%% file: acl2020.tex
\documentclass[11pt,a4paper]{article}
\usepackage[hyperref]{emnlp2020}
\usepackage{times}
\usepackage{latexsym}
\usepackage{graphicx}
\usepackage{multirow}
\usepackage{url}
\usepackage{amssymb}
\usepackage{amsmath}
\usepackage{enumitem}
\usepackage{color,soul}
\usepackage{subfig}

\usepackage{microtype}

\aclfinalcopy 
\def\aclpaperid{3315} 

\title{Improving Limited Labeled Dialogue State Tracking with Self-Supervision}

\author{Chien-Sheng Wu, Steven Hoi, and Caiming Xiong \\
Salesforce Research \\
\texttt{[wu.jason, shoi, cxiong]@salesforce.com}
}

\date{}

\begin{document}
\maketitle
\begin{abstract}
Existing dialogue state tracking (DST) models require plenty of labeled data. However, collecting high-quality labels is costly, especially when the number of domains increases.
In this paper, we address a practical DST problem that is rarely discussed, i.e., learning efficiently with limited labeled data.
We present and investigate two self-supervised objectives: preserving latent consistency and modeling conversational behavior. 
We encourage a DST model to have consistent latent distributions given a perturbed input, making it more robust to an unseen scenario.
We also add an auxiliary utterance generation task, modeling a potential correlation between conversational behavior and dialogue states.
The experimental results show that our proposed self-supervised signals can improve joint goal accuracy by 8.95\% when only 1\% labeled data is used on the MultiWOZ dataset.
We can achieve an additional 1.76\% improvement if some unlabeled data is jointly trained as semi-supervised learning.
We analyze and visualize how our proposed self-supervised signals help the DST task and hope to stimulate future data-efficient DST research.

\end{abstract}

\section{Introduction}
Dialogue state tracking is an essential component in task-oriented dialogue systems designed to extract user goals/intentions expressed during a conversation. Accurate DST performance can facilitate downstream applications such as dialogue management.
However, collecting dialogue state labels is very expensive and time-consuming~\cite{multiwoz}, requiring dialogue experts or trained turkers to indicate all \textit{(domain, slot, value)} information for each turn in dialogues. 
This problem becomes important from single-domain to multi-domain scenarios. It will be more severe for a massive-multi-domain setting, making DST models less scalable to a new domain.

Existing DST models require plenty of state labels, especially those ontology-based DST approaches~\cite{henderson2014word, NBT, GLAD}.  
They assume a predefined ontology that lists all possible values is available, but an ontology requires complete state annotation and is hard to get in real scenario~\cite{P18-1134PtrNet}.
They also cannot track unseen slot values that are not predefined.
Ontology-free approaches~\cite{P18-1134PtrNet, chao2019bert}, on the other hand, are proposed to generate slot values from dialogue history directly. 
They achieve good performance on multi-domain DST by copy-attention mechanism but still observe a significant performance drop under limited labeled data scenario~\cite{wu-2019-trade}. 

In this paper, we approach the DST problem using copy-augmented ontology-free models from a rarely discussed perspective, assuming that only a few dialogues in a dataset have annotated state labels.
We present two self-supervised learning (SSL) solutions: 
1) Preserving latent consistency: We encourage a DST model to have similar latent distributions (e.g., attention weights and hidden states) for a set of slightly perturbed inputs. This assumption is known as consistency assumption~\cite{zhou2004learning, chapelle2009semi,berthelot2019mixmatch} in semi-supervised learning, making distributions sufficiently smooth for the intrinsic structure collectively. 
2) Modeling conversational behavior: We train a DST model to generate user utterances and system responses, hoping that this auxiliary generation task can capture intrinsic dialogue structure information and benefit the DST performance. This training only needs dialogue transcripts and does not require any further annotation. We hypothesize that modeling this potential correlation between utterances and states is helpful for generalization, making a DST model more robust to unseen scenarios.


We simulate limited labeled data using MultiWOZ~\cite{multiwoz}, one of the task-oriented dialogue benchmark datasets, with 1\%, 5\%, 10\%, and 25\% labeled data scenarios.
The experimental results of 1\% data setting show that we can improve joint goal accuracy by 4.5\% with the proposed consistency objective and with an additional 4.43\% improvement if we add the behavior modeling objective.
Furthermore, we found that a DST model can also benefit from those remaining unlabeled data if we joint train with their self-supervised signals, suggesting a promising research direction of semi-supervised learning.
Lastly, we visualize the learned latent variables and conduct an ablation study to analyze our approaches. 

\section{Background}
Let us define $X_{1:T} = \{(U_1, R_1), \dots, (U_T, R_T)\}$ as the set of user utterance and system response pairs in $T$ turns of a dialogue, and $B = \{B_1,\dots,B_T\}$ are the annotated dialogue states. Each $B_t$ contains a set of \textit{(domain, slot, value)} tuples accumulated from turn $1$ to turn $t$, therefore, the number of tuples usually grows with turn $t$. 
Note that it is possible to have multiple domains triggered in the same state $B_t$. A dialogue example and its labeled states are shown in Table~\ref{tb:example}.

\begin{table}
\centering
\resizebox{\linewidth}{!}{
\begin{tabular}{|r|l|l|l|}
\hline
Usr & \multicolumn{3}{l|}{Can you help me find a \textbf{nightclub} in \textbf{south} Cambridge?} \\ \hline
Sys & \multicolumn{3}{l|}{\begin{tabular}[c]{@{}l@{}}\textbf{The Night} is located at 22 Sidney St. Their phone number \\ is 01223324600. You will need to call for their entry fee.\end{tabular}} \\ \hline
Usr & \multicolumn{3}{l|}{Can you schedule me a taxi to take me there?} \\ \hline
Sys & \multicolumn{3}{l|}{\begin{tabular}[c]{@{}l@{}} Can book you a taxi. Can you tell me the arrival or \\ departure time ?\end{tabular}} \\ \hline
Usr & \multicolumn{3}{l|}{Also, I need a hotel with \textbf{parking} and \textbf{2 stars}.} \\ \hline
\multicolumn{2}{|r|}{Annotated State} 
& \multicolumn{2}{l|}{\begin{tabular}[c]{@{}l@{}} (attraction, type, nightclub), \\ (attraction, area, south), \\ (attraction, name, The Night), \\ (hotel, parking, yes), (hotel, stars, 2) \end{tabular}} \\ \hline
\end{tabular}
}
\caption{A multi-domain dialogue example in MultiWOZ.}
\label{tb:example}
\end{table}

\begin{figure}[t]
\centering
\includegraphics[width=\linewidth]{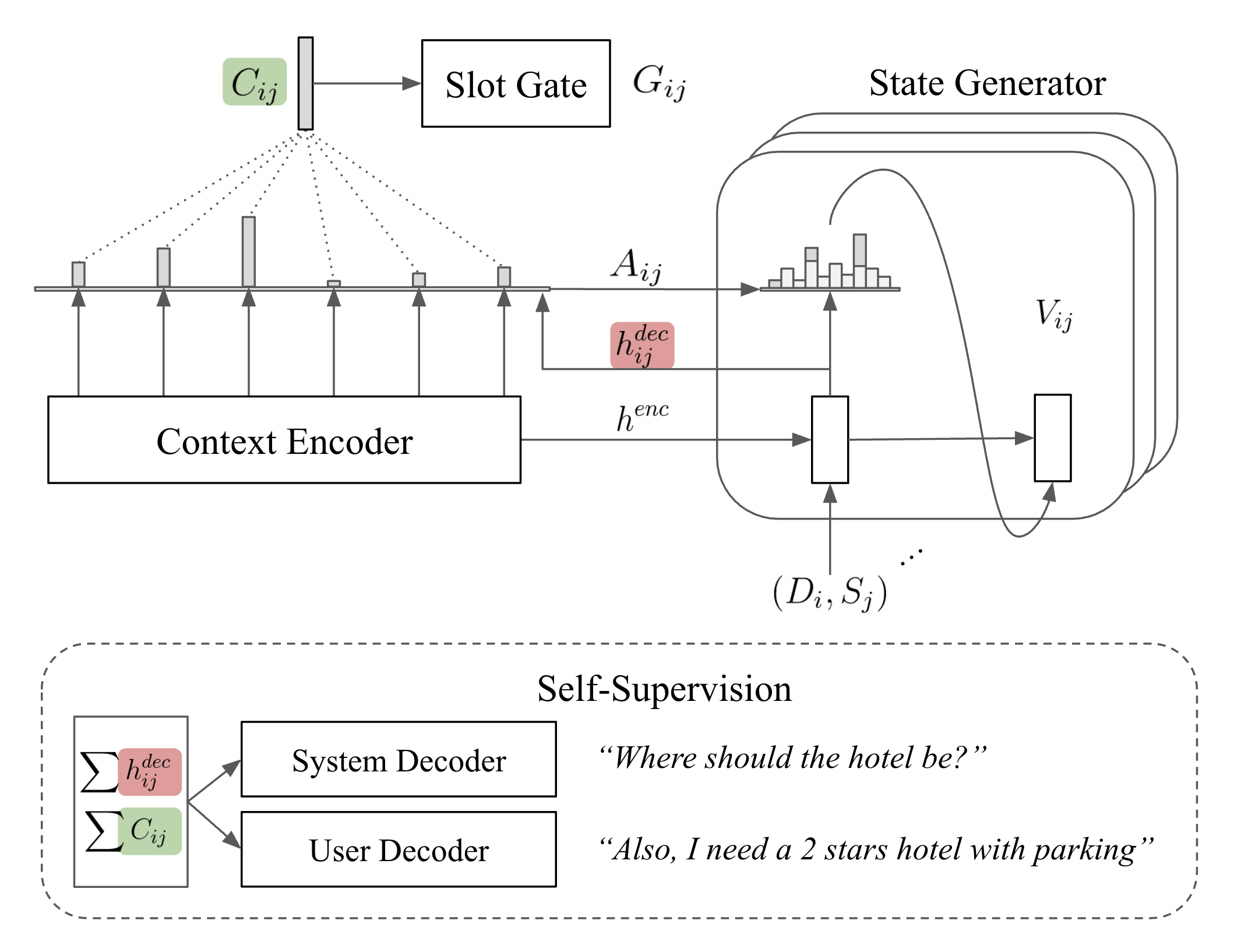}
\caption{The block diagram of copy-attention ontology-free framework for dialogue state tracking. The self-supervised modules (dotted parts) are discarded during inference time.}
\label{model}
\end{figure}

We briefly introduce a common approach for ontology-free DST in the following.
As shown in Figure~\ref{model}, a context encoder encodes dialogue history $X_{1:t}$, and a state generator decodes slot values $V_{ij}$ for each \textit{(domain, slot)} pair $\{(D_i, S_j)\}$, where $i$ denotes the domain index and $j$ is the slot index. 
The context encoder and the state generator can be either a pre-trained language model or a simple recurrent neural network.
During the decoding stage for each $V_{ij}$, a copy-attention mechanism such as text span extraction~\cite{vinyals2015pointer} or pointer generator~\cite{see2017PG,wu2019global} approach is added to the state generator and strengthen its value generation process. 

Moreover, many ontology-free DST models are also equipped with a slot gate mechanism \cite{P18-1134PtrNet,rastogi2019towards,zhang2019find}, which is a classifier that predicts whether a \textit{(domain, slot)} pair is mentioned, not mentioned, or a user does not care about it.
In this pipeline setting, they can add additional supervision to their models and ignore the not mentioned pairs' prediction. 
More specifically, the \textit{(domain, slot)} pair $\{(D_i, S_j)\}$ obtains its context vector $C_{ij}$ to predict a slot gate distribution $G_{ij}$. The context vector $C_{ij}$ is the weighted-sum of encoder hidden states using the attention distribution $A_{ij}$, and $G_{ij}$ is a three-way classification distribution mapping from the context vector:
\begin{equation}
    \begin{array}{c}
        G_{ij} = \textnormal{FFN}(C_{ij}) \in \mathbb{R}^{3}, \\
        C_{ij} = A_{ij} h^{enc} \in \mathbb{R}^{d_{emb}}, \\
        A_{ij} = \textnormal{Softmax}(\textnormal{Dist}(h^{dec}_{ij}, h^{enc})) \in \mathbb{R}^{M}, \\
    \end{array}
\end{equation}
where $d_{emb}$ is the hidden size, $h^{enc} \in \mathbb{R}^{M \times d_{emb}}$ is hidden states of the context encoder for $M$ input words, and $h^{dec}_{ij} \in \mathbb{R}^{d_{emb}}$ is the first hidden state of the state generator. The Dist function can be any vector similarity metric, and FFN can be any kind of classifier.


Such model is usually trained end-to-end with two loss functions, one for slot values generation and the other for slot gate prediction. The overall supervised learning objective from the annotated state labels is 
\begin{equation}
    L_{sl} = \sum^{|ij|} H(V_{ij}, \hat{V}_{ij}) + H(G_{ij}, \hat{G}_{ij}),
\end{equation}
where $H$ is the cross-entropy function. The total number of \textit{(domain, slot)} pairs is $|ij|$, and there are 30 pairs in MultiWOZ.



\section{Self-Supervised Approaches}
This section introduces how to leverage dialogue history $ X $, which is easy to collect, to boost DST performance without annotated dialogue state labels implicitly. 
We first show how we preserve latent consistency using stochastic word dropout, and we discuss our design for utterance generation.

\subsection{Latent Consistency}
\begin{figure}[t]
\centering
\includegraphics[width=\linewidth]{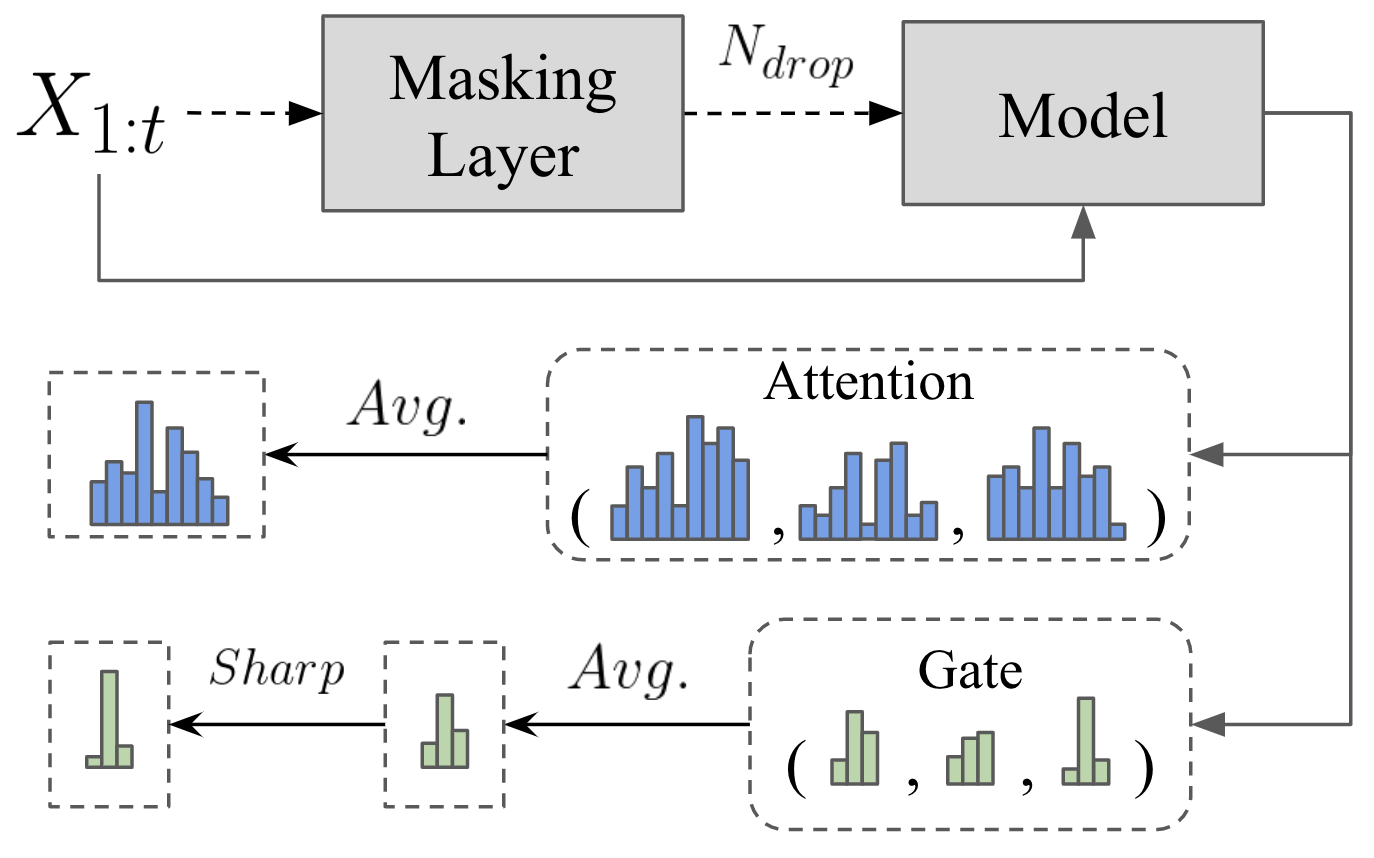}
\caption{The block diagram of preserving latent consistency. $N_{drop} + 1$ attention and slot gate distributions are averaged (and sharpened) to be the guessed distribution.}
\label{model_consist}
\end{figure}

The goal of preserving latent consistency is that DST models should be robust to a small perturbation of input dialogue history. As shown in Figure~\ref{model_consist}, we first randomly mask out a small number of input words into unknown words for $N_{drop}$ times. Then we use $N_{drop}$ dialogue history together with the one without dropping any word as input to the base model and obtain $N_{drop}+1$ model predictions.

Masking words into unknown words can also strengthen the representation learning because when important words are masked, a model needs to rely on its contextual information to obtain a meaningful representation for the masked word. For example, ``I want a cheap restaurant that does not spend much.'' becomes ``I want a [UNK] restaurant that [UNK] not spend much.'' This idea is motivated by the masked language model learning~\cite{devlin-etal-2019-bert}. We randomly mask words instead of only hiding slot values because it is not easy to recognize the slot values without ontology. 

Afterward, we produce a ``gues'' for its latent variables: the attention distribution and the slot gate distribution in our setting. 
Using the $N_{drop}+1$ model’s predictions, we follow the label guessing process in MixMatch algorithm~\cite{berthelot2019mixmatch} to obtain a smooth latent distribution.
We compute the average of the model’s predicted distributions by
\begin{equation}
    \begin{split}
        \hat{A}_{ij}^{*}, \hat{G}_{ij}^{*} = \frac{\sum\limits_{d=1}^{N_{drop}+1} P(A_{ij}, G_{ij} |X_{1:t}^d, \theta)}{N_{drop}+1}, 
    \end{split}
\end{equation}
where $\theta$ is the model parameters. 
$A_{ij}$ and $G_{ij}$ are the smooth latent distribution that we would like a DST model to follow.
We include the original input without word masking input the average. During the early stage of training, we may not have a good latent distribution even if it has labeled supervision. 

Furthermore, inspired by the common usage of entropy minimization~\cite{grandvalet2005semi}, we perform one more step for the gate distribution. We apply a sharpening function, adjusting the temperature $T$ of the categorical distribution, to reduce the entropy of slot gate prediction.
\begin{equation}
    \begin{array}{c}
        \hat{G}_{ij}^{**} = \textnormal{Sharp}(\hat{G}_{ij}^{*}, T), \\
        \textnormal{Sharp}(p, T)_i = p_i^{\frac{1}{T}} / \sum  p_i^{\frac{1}{T}}.
    \end{array}
\end{equation}
In this way, we encourage a DST model to be more confident to its gate prediction as $T$ decreases, since the sharpen $\hat{G}_{ij}^{**}$ will approach a one-hot distribution when $T = 0$.
The sharpening function is not applied to the predicted attention distribution because we do not expect and force attention distribution to be a sharp categorical distribution. 

We use the two guessed distributions to train a DST model to be consistent for the attention and slot gate given noise inputs. The following consistency loss is added:
\begin{equation}
    \begin{split}
       L_{cons} = \sum\limits^{|ij|} \sum\limits^{N_{drop}+1}_d (
       & \textnormal{MSE}(\hat{G}_{ij}^{**}, \hat{G}_{ij}^d)  \\
       & + \textnormal{MSE}(\hat{A}_{ij}^{*}, \hat{A}_{ij}^d) ). 
    \end{split}
\end{equation}
We follow \citet{berthelot2019mixmatch} to apply the mean-squared error function as our loss function.

We train a model to be consistent in terms of latent distributions because it is hard to guarantee the quality of generated values in different perturbed input, especially when we do not have much labeled data. Also, each perturbed sample may generate slot values that have different number of words, and maintaining consistency of sequential distributions could be challenging. As a result, we use slot gate distribution and attention distribution as intermediate targets since the former is the first stage for the whole prediction process, and the latter directly influences the copy mechanism.

\subsection{Conversational Behavior Modeling}
We hypothesize that with similar dialogue states, a system will reply also similar responses. For example, when a system asks ``What is your taxi destination from Palo Alto?'', then we can infer that system's state may include \textit{(taxi, departure, Palo Alto)}. In this way, we can potentially model the correlation between dialogue states and dialogue behavior. In practice, we use two decoders, one modeling user and one modeling system behavior, to generate utterances based on the learned representations from a DST model.

We use a gated recurrent unit (GRU) to generate the next system response based on the dialogue history $X_{1:t}$ and current predicted dialogue states $B_t$, and use another GRU to generate/recover user utterance based on last dialogue history $X_{1:t-1}$ and current predicted dialogue states $B_t$. Intuitively, we expect the system GRU to capture correlation between $R_{t+1}$ and $B_{t}$, and the user GRU to learn for $U_t$ and $B_t$.
GRUs generate a sequence of words during training and compute cross-entropy losses between generated sentences and target sentences. We do not use the attention mechanism intentionally because 1) our goal is not to have an outstanding performance on sentence generation, and 2) we expect the model can generate sentences by solely aligning its initial states from a DST model.

As shown in Figure~\ref{model}, we initial our system and user GRUs using latent variables from an ontology-free DST model. The initial state $h_{init}$ to be aligned is defined by
\begin{equation}
    h_{init} = \sum^{|ij|} [h^{dec}_{ij}; C_{ij}],
    \label{eq:cij}
\end{equation}
where $[;]$ denotes vector concatenation and we sum representations from all \textit{(domain, slot)} pairs. We use the context vector $C_{ij}$ to represent dialogue history, and $h^{dec}_{ij}$ to represent dialogue state. 
The overall self-supervised loss function for modeling conversational behavior is 
\begin{equation}
    L_{cb} = H(R_{t+1}, \hat{R}_{t+1}) + H(U_t, \hat{U}_t),
\end{equation}
where $\hat{R}_{t+1}$ and $\hat{U}_t$ are predicted response and user utterance initialized by the $h_{init}$ vector.

\subsection{Overall Objectives}
During training, we optimize both supervised signal and self-supervised signal using the labeled data. The overall loss function is
\begin{equation}
    \begin{array}{c}
        L_{label} = L_{sl} + \alpha L_{cb} + \beta L_{cons},
    \end{array}
\end{equation}
where $\alpha$ and $\beta$ are hyper-parameters. 

Other than labeled data, we can also sample unlabeled data to perform self-supervision as a regularization term. This strategy can be considered as a semi-supervised approach, leveraging unlabeled data to learn a smooth prediction. For unlabeled data, we use only the self-supervised signal to update the model,
\begin{equation}
    \begin{array}{c}
        L_{unlabel} = L_{cb} + \beta L_{cons}.
    \end{array}
\end{equation}
In practice, we first draw a batch of samples from labeled data to update the model's parameters and then draw another batch of samples from unlabeled data. We find that taking turns to train unlabeled data with labeled data works better than pre-training with unlabeled data then fine-tuning on labeled data.

\begin{table*}
\centering
\resizebox{0.95\linewidth}{!}{
\begin{tabular}{r|r|c|c|c|c}
\hline
\multicolumn{2}{r|}{} & \textbf{1\%} & \textbf{5\%} & \textbf{10\%} & \textbf{25\%} \\ \hline
\multicolumn{2}{r|}{\textit{TRADE (w/o Ont.)}~\cite{wu-2019-trade}} & 9.70 (11.74) & 29.38 (32.41) & 34.07 (37.42) & 41.41 (44.01) \\ 
\hline
\multicolumn{2}{r|}{+ Consistency}  & 14.22 (15.77) & 30.18 (33.59) & 36.14 (39.03) & 41.38 (44.33) \\ 
\hline
\multicolumn{2}{r|}{+ Behavior}  & 18.31 (20.59) & 31.13 (34.38) & 36.90 (40.70) & 42.48 (45.12) \\ 
\hline
\multicolumn{2}{r|}{Consistency + Behavior}  & 18.65 (21.21) & 31.61 (35.67) & 37.05 (40.29) & \textbf{42.71 (45.21)} \\ 
\hline
\multicolumn{2}{r|}{\begin{tabular}[r]{@{}r@{}}Consistency + Behavior \\+ Unlabeled Data\end{tabular}} & \textbf{20.41 (23.0)} & \textbf{33.67 (37.82)} & \textbf{37.16 (40.65)} & 42.69 (45.14) \\ 
\hline
\hline
\multicolumn{2}{r|}{\textit{SUMBT (w/ Ont.)}~\cite{lee-2019-sumbt}} & 4.30 (-) & 30.56 (-) & 38.31 (-) & 42.59 (-) \\ 
\hline 
\multicolumn{2}{r|}{\textit{TOD-BERT (w/ Ont.)}~\cite{wu2020tod}} & 10.3 (-) & 27.8 (-) & 38.8 (-) & 44.3 (-) \\ 
\hline 
\multicolumn{2}{r|}{\textit{DSDST-Span (w/o Ont.)}~\cite{zhang2019find}} & 19.82 (-) & 32.20 (-) & 37.81 (-) & 39.48 (-) \\ 
\hline 
\end{tabular}
}
\caption{Joint goal accuracy and its fuzzy matching version in parentheses on MultiWOZ test set from 1\% to 25\% labeled training data. As a reference, we test some other DST trackers that using the pre-trained language model BERT~\cite{devlin-etal-2019-bert} under limited labeled scenario, as shown in the last few rows.}
\label{tb:trade_result_self}
\end{table*}

\section{Experiments}

\subsection{Base Model}
In this paper, we focus on applying self-supervision for ontology-free DST approaches. 
We select TRADE~\cite{wu-2019-trade} model as the base model. 
We select TRADE because
1) it is a pointer-generator based dialogue state tracker with a copy-attention mechanism that can generate unseen slot values, and
2) it is one of the best ontology-free models that show good domain generalization ability in its zero-shot and few-shot experiments, and it is open-source \footnote{\url{github.com/jasonwu0731/trade-dst}}.
Note that our proposed self-supervised training objectives are not limited to one DST model. For example, the BERTQA-based span extraction methods~\cite{chao2019bert, gao2019dialog} can be applied with slight modification, viewing [CLS] token as the encoded vector and the span distributions as the slot contextual representations. 

\begin{table}
\centering
\resizebox{0.9\linewidth}{!}{
\begin{tabular}{r|c|c|c|c|l}
\hline
\multicolumn{1}{l|}{} & \textbf{1\%} & \textbf{5\%} & \textbf{10\%} & \textbf{25\%} & \textbf{100\%} \\ \hline
Hotel & 33 & 174 & 341 & 862 & 3381 \\
Train & 35 & 166 & 332 & 809 & 3103 \\
Attraction & 29 & 143 & 276 & 696 & 2717 \\
Restaurant & 36 & 181 & 377 & 928 & 3813 \\
Taxi & 11 & 71 & 150 & 395 & 1654 \\ \hline
Total* & 84 & 421 & 842 & 2105 & \multicolumn{1}{c}{8420} \\ \hline
\end{tabular}
}
\caption{Number of simulated labeled dialogues on MultiWOZ training set. (* Total number of dialogues is less than the summation of dialogues in each domain because each dialogue has multiple domains.)}
\label{tb:number_of_dial}
\end{table}



\subsection{Dataset}
MultiWOZ~\cite{multiwoz} is one of the largest existing human-human conversational corpus spanning over seven domains, containing around 8400 multi-turn dialogues, with each dialogue averaging 13.7 turns. We follow~\citet{wu-2019-trade} to only use the five domains (\textit{hotel}, \textit{train}, \textit{attraction}, \textit{restaurant}, \textit{taxi}) because the other two domains (\textit{hospital}, \textit{police}) have very few dialogues (10\% compared to others) and only exist in the training set. In total, there are 30 \textit{(domain, slot)} pairs. We also evaluate on its revised version 2.1 from \citet{eric2019multiwoz} in our experiments, due to the space limit, results on version 2.1 are reported in the Appendix.

We simulate a limited labeled data scenario by randomly selecting dialogues from the original corpus using a fixed random seed. The dataset statistics of each labeled ratio is shown in Table~\ref{tb:number_of_dial}. For example, in 1\% labeled data setting, there are 84 dialogues across five different domains. Note that the summation of dialogues from each domain is more than the number of total dialogues because each dialogue could have more than one domain, e.g., two domains are triggered in the Table~\ref{tb:example}.

\subsection{Training Details}
The model is trained end-to-end using Adam optimizer~\citep{KingmaB14} with a batch size of 8 or 32. A grid search is applied for $\alpha$ and $\beta$ in the range of 0.1 to 1, and we find that models are sensitive to different $\alpha$ and $\beta$. The learning rate annealing is used with a 0.2 dropout ratio. All the word embeddings have 400 dimensions by concatenating 300 Glove embeddings \cite{pennington2014glove} and 100 character embeddings \cite{hashimoto2016joint}. A greedy decoding strategy is used for the state generator because the slot values are usually short in length. We mask out 20\%-50\% of input tokens to strengthen prediction consistency. The temperature $T$ for sharpening is set to 0.5, and augmentation number $N_{drop}$ is 4.

\subsection{Results}
\label{sec:results}
Joint goal accuracy and its fuzzy matching~\footnote{\url{github.com/seatgeek/fuzzywuzzy}} version are used to evaluate the performance on multi-domain DST. The joint goal accuracy compares the predicted dialogue states to the ground truth $B_t$ at each dialogue turn $t$, and the output is considered correct if and only if all the \textit{(domain, slot, value)} tuples exactly match the ground truth values in $B_t$, which is a very strict metric. The fuzzy joint goal accuracy is used to reward partial matches with the ground truth~\cite{rastogi2019towards}. For example, two similar values ``Palo Alto'' and ``Palo Alto city'' have a fuzzy score of 0.78.

In Table~\ref{tb:trade_result_self}, we evaluate four different limited labeled data scenarios: 1\%, 5\%, 10\%, and 25\%. We test our proposed self-supervised signals by only adding latent consistency objective (row 2), only adding conversational behavior objective (row 3), using both of them (row 4), and using both of them together with unlabeled data (row 5). In general, we find that each self-supervision signal we presented is useful in its degree, especially for 1\% and 5\% labeled data scenarios. Modeling conversational behavior seems to be more effective than preserving prediction consistency, which is not surprising because the latter is a point-wise self-supervised objective function. We also found that self-supervision becomes less dominant and less effective as the number of labeled data increases. We try 100\% labeled data with self-supervision, and it only achieves slight improvement, 48.72\% joint goal accuracy compared to the original reported 48.62\%.

\begin{table}[t]
\centering
\resizebox{\linewidth}{!}{
\begin{tabular}{r|c|c}
\hline
\multicolumn{1}{c|}{} & Gate Acc ($\uparrow$) & Attention KL ($\downarrow$) \\ \hline
100\% Data & 97.61 & - \\ \hline
1\% Data w/o SSL & 91.38 & 10.58 \\
1\% Data w/ SSL & \textbf{94.30} & \textbf{6.19} \\ \hline
\end{tabular}
}
\caption{Gate accuracy on 1\% data improves 2.92\% and KL divergence between 1\% and 100\% data decreases 4.39 with self-supervision.}
\label{tb:analysis}
\end{table}

Taking a closer look to the results in Table~\ref{tb:trade_result_self}, preserving consistency has 4.52\% (or 4.03\% fuzzy) improvement for 1\% scenario. Once the labeled data increases to 25\% (2105 dialogues), there is no difference with or without the consistency objective. Meanwhile, modeling conversational behavior objective seems to be more effective than the consistency objective, as it has 8.61\% (or 8.85\% fuzzy) improvement. A small improvement can be further observed if we combine both of them and jointly train end-to-end. When we also leverage those remaining dialogue data and conduct semi-supervised learning, we can achieve the highest joint goal accuracy, 20.41\% in 1\% setting, and 33.67\% in 5\% setting. In these experiments, we simply use the remaining dialogues in the dataset as unlabeled data, e.g., 1\% labeled with 99\% unlabeled, 5\% labeled with 95\% unlabeled, etc.

We also test some other DST trackers in the last few rows in Table~\ref{tb:trade_result_self}, which all of them are replied on the pre-trained language model BERT~\cite{devlin-etal-2019-bert}. SUMBT~\cite{lee-2019-sumbt} and TOD-BERT~\cite{wu2020tod} are ontology-based approaches. The former uses BERT to encode each utterance and builds an RNN tracker on top of BERT. The latter uses its pre-trained task-oriented dialogue BERT to encode dialogue history and adds simple slot-dependent classifiers.
Note that we still assume they have a full ontology in this setting even though it is not a fair comparison under a limited labeled scenario. DSDST-Span~\cite{zhang2019find} is an ontology-free DST tracker, it uses BERT to encode dialogue history together with each \textit{(domain, slot)} pair separately and extract a corresponding text span as its slot values.

\section{Analysis and Visualization}
We would interpret how self-supervised signals help to learn better DST performance.
The first interesting observation is that the key improvement comes from the slot-dependent context vectors $C_{ij}$. If we remove the context vector $C_{ij}$ from Eq (\ref{eq:cij}), the performance of 1\% labeled data setting drops from 18.31\% to 11.07\%. The next question is: what do these contextual vectors influence? First, context vectors are the weighted-sum of encoder hidden states, which means they correlate with the learned attention distribution. Also, context vectors are used to predict slot gates, which is essential to be able to trigger the state generator. Therefore, using self-supervision to align contextual slot vectors may help get better attention distributions and better slot gate prediction.

\paragraph{Slot Gate}
As shown in Table~\ref{tb:analysis}, gate accuracy of 1\% labeled data improves by around 3\% with self-supervision. We also compare attention distributions among a model trained with 1\% labeled data, a model trained with 1\% labeled data and self-supervision, and a model trained with 100\% labeled data. We observe a smaller value of KL divergence with self-supervision (the lower, the better), i.e., the attention distribution becomes more similar to the one learned from 100\% labeled data, which we assume that it is supposed to be a better attention distribution.

\begin{figure}[t]
\centering
\includegraphics[width=\linewidth]{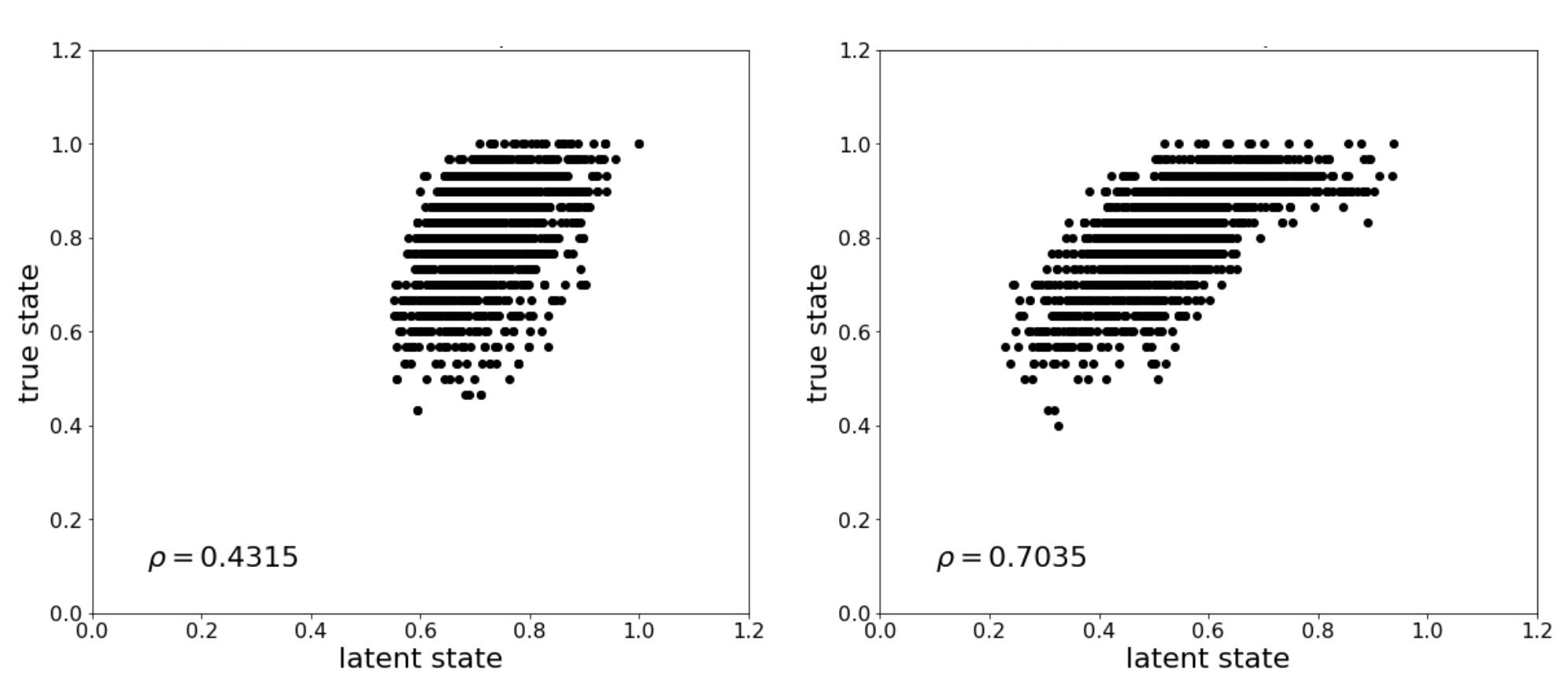}
\caption{The correlation on test set between latent dialogue states and true dialogue states on 1\% labeled data. Left-hand side is without self-supervision and right-hand side is with self-supervision.}
\label{state_sim}
\end{figure}

\begin{figure}[t]
\centering
\resizebox{0.90\linewidth}{!}{
\begin{tabular}{|c|c|}
\hline
 & Dialogue History \\ \hline
\rotatebox[origin=l]{90}{100\% Data} & \includegraphics[width=\linewidth]{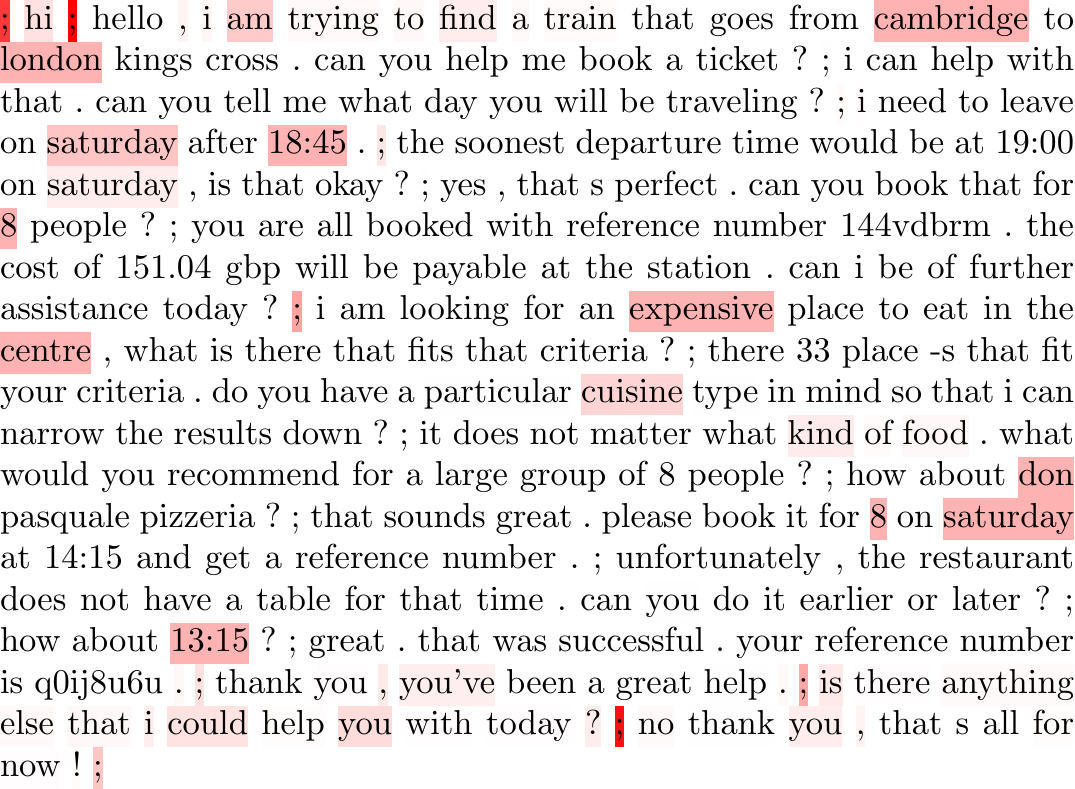} \\ \hline
\rotatebox[origin=l]{90}{1\% Data w/o Self-supervision} & \includegraphics[width=\linewidth]{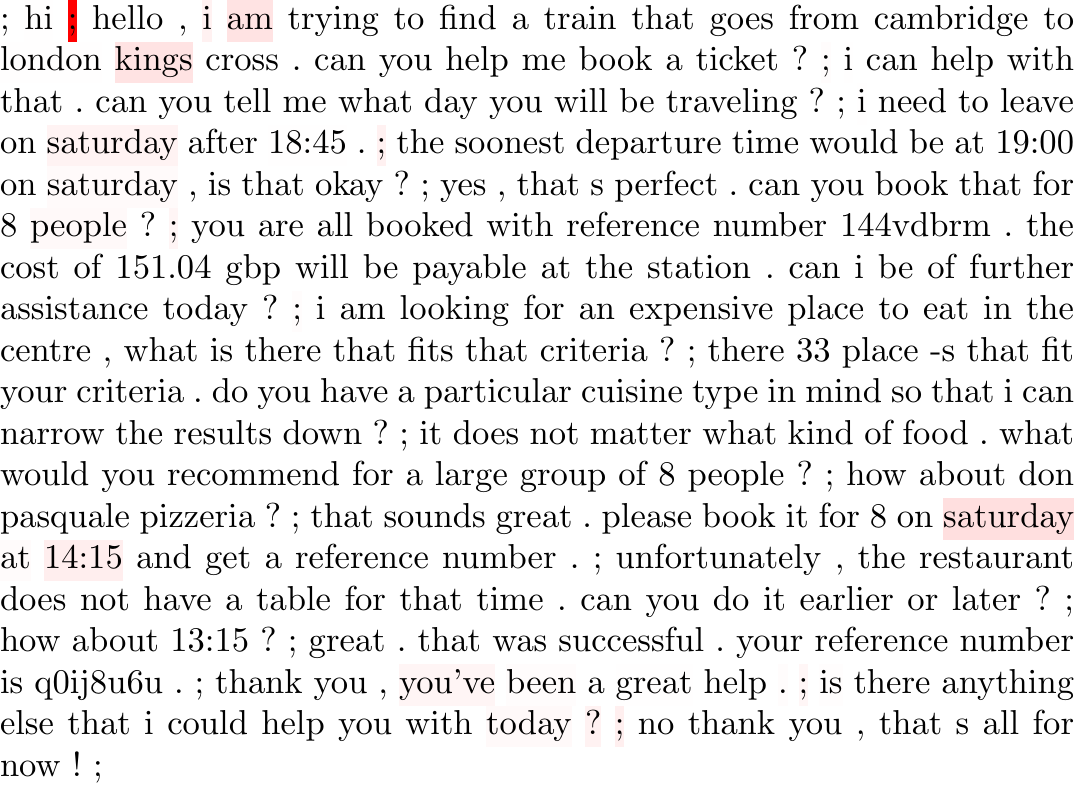} \\ \hline
\rotatebox[origin=l]{90}{1\% Data w/ Self-supervision}  & \includegraphics[width=\linewidth]{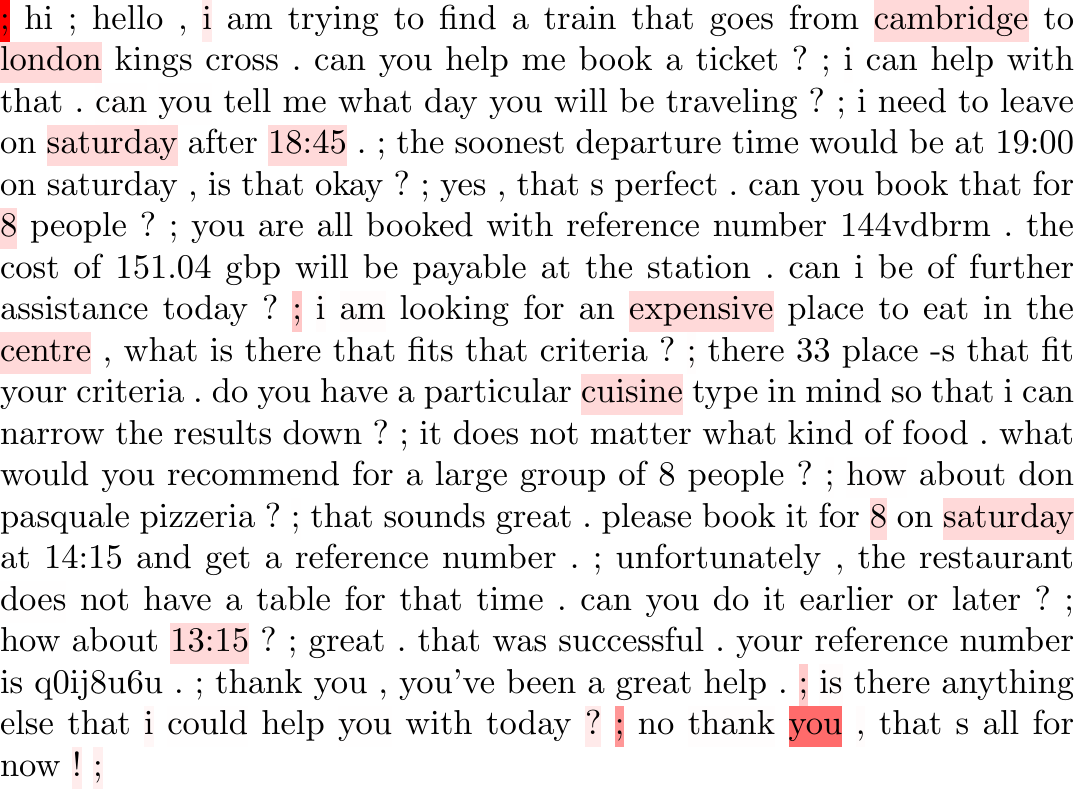} \\ \hline
\end{tabular}
}
\caption{Attention visualization for a dialogue history. The darker color means higher attention weight. The 1\% labeled data model with self-supervision learns attention distribution more similar to the one using 100\% labeled data.}
\label{attn_plot_1}
\end{figure}

We randomly pick up 2,000 dialogue turns on the test set to compute the correlation between latent learned states ($h_{init}$) of 1\% labeled data and the true gating status ($G$) of the \textit{(domain, slot)} pairs.
As shown in Figure~\ref{state_sim}, the x-axis is the cosine similarity score between two latent dialogue states the model learned, and the y-axis is the cosine similarity score of their true gating status. 
Ideally, when the slot gate status is similar, then the learned representations should also have a high similarity score.
We find the model trained with self-supervision (right) has a higher Pearson correlation coefficient than the one without (left), increasing from 0.4315 to 0.7035, implying that with self-supervision, models can learn better state representations. 

\paragraph{Copy Attention}
We also visualize the attention distributions of a dialogue history in Figure~\ref{attn_plot_1}. The darker red color means the higher attention weight and the higher copy probability. We sum attention distributions of $A_{ij}$ for all \textit{(domain, slot)} pairs and normalize it. The 1\% labeled data model with self-supervision has an attention distribution similar to the one using 100\% labeled data. For example, both of them focus on some useful slot information such as ``Cambridge'', ``London'', ``Saturday'', and ``18:45''. The results of attention distribution are crucial, especially in our limited labeled setting. The higher the attention weight, the higher the probability that such word will be copied from the dialogue history to the output slot values. More attention visualizations are shown in the Appendix.

\begin{figure*}[t]
\centering
\includegraphics[width=0.65\linewidth]{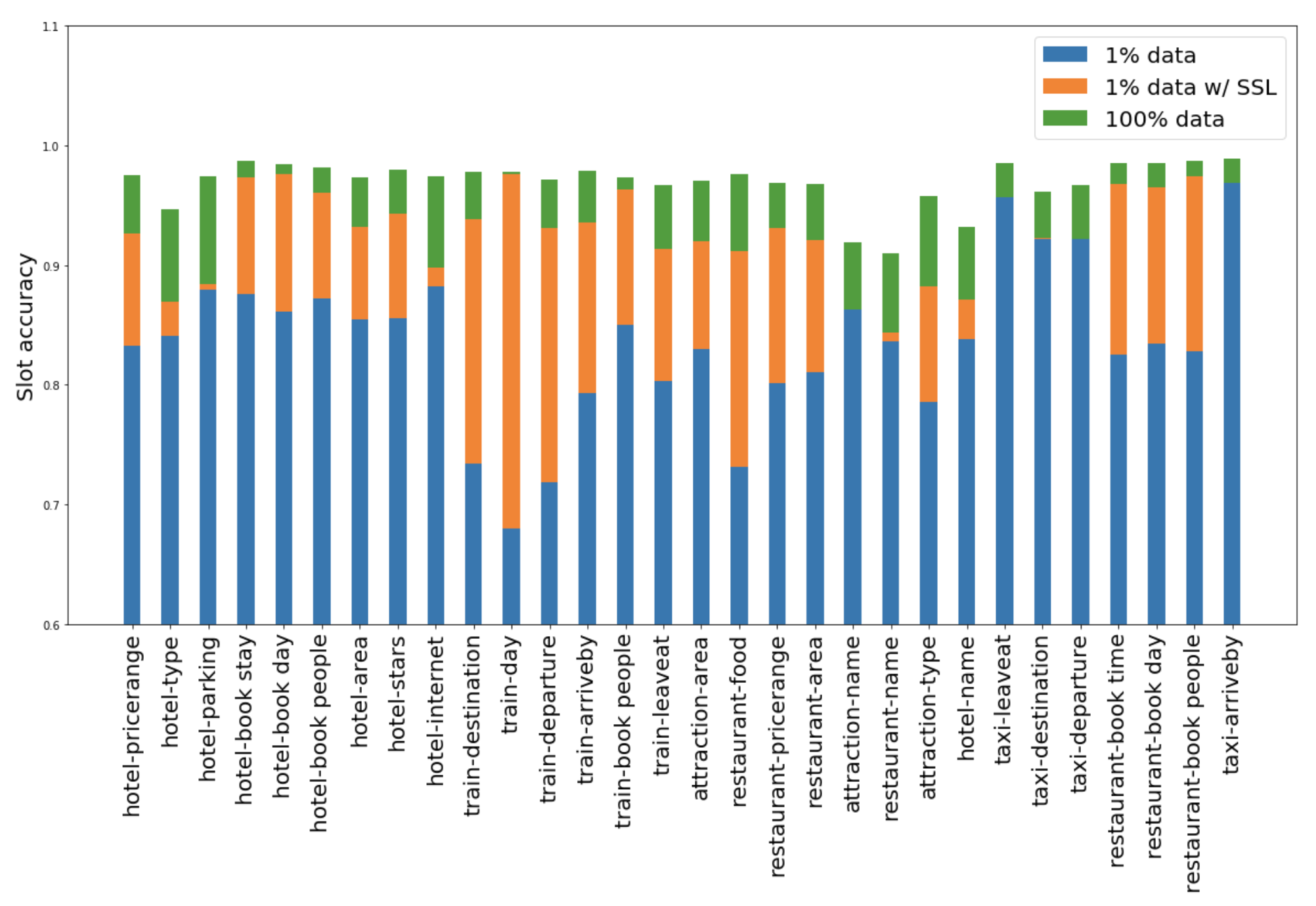}
\caption{Slot accuracy visualization for each \textit{(domain, slot)} pairs. Several slots such as \textit{(train, day)} and \textit{(hotel, book stay)} that using 1\% data with self-supervision almost perform the same as using 100\% data.}
\label{slot_acc}
\end{figure*}

\paragraph{Slot Accuracy Analysis}
We are interested in which domains and which slots are easier to be self-supervised learned. As shown in Figure~\ref{slot_acc}, the x-axis is each \textit{(domain, slot)} pair, and the y-axis is its slot accuracy (at each dialogue turn whether the pair is predicted correctly). The blue bar is the performance of 1\% labeled data without self-supervision. The orange part is the improvement by using self-supervision. The green part can be viewed as the upper-bound of the base model using 100\% labeled data.

The top three \textit{(domain, slot)} pairs that is most effective with self-supervision are \textit{(train, day)}, and \textit{(train, departure)}, \textit{(train, destination)}. On the other hand, self-supervision are less helpful to pairs such as \textit{(hotel, parking)}, \textit{(hotel, internet)}, \textit{(restaurant, name)}, and all the pairs in the \textit{taxi} domain. One possible reason is that self-supervision is sensitive to the unlabeled data size, i.e., the major domain is dominant in the overall performance. It is worth mentioning that in the \textit{taxi} domain, all the slots perform relatively well with 1\% labeled data. This could also explain why the zero-shot performance reported in \citet{wu-2019-trade} is much better than the other four domains.


\section{Related Work}
\paragraph{Dialogue State Tracking}
Traditional dialogue state tracking models combine semantics extracted by language understanding modules to estimate the current dialogue states~\cite{williams2007partially, thomson2010bayesian, wang2013simple, williams2014web}, or to jointly learn speech understanding~\cite{henderson2014word, zilka2015incremental}. One drawback is that they rely on hand-crafted features and complex domain-specific lexicons besides the ontology, and are difficult to extend and scale to new domains. As the need for domain expanding, research direction moves from single domain DST setting and datasets~\cite{Wen-WOZ} to multi-domain DST setting and datasets~\cite{multiwoz, eric2019multiwoz}.

There are three main categories to perform DST, ontology-based, partial-ontology-based, and ontology-free approaches. Ontology-based methods~\cite{NBT, Wen-WOZ, rastogi2017scalable, ren2018towards, GLAD, MDBT, lee-2019-sumbt,chenschema} train metric learning functions for context encoder and ontology encoder, and score over a predefined slot value candidates. Partial-ontology-based~\cite{goel2019hyst, zhang2019find, rastogi2019towards} approaches only use part of an ontology to perform ranking and use generation techniques for the remaining slots. Ontology-free methods~\cite{chao2019bert, gao2019dialog, ren2019scalable, kumar2020ma,wu-2019-trade,kumar2020ma,kim2019efficient} rely on generation with copy mechanism without predefined ontology, which has better generalization ability to unseen slot values. Our work is closer to ontology-free approaches because it is reasonable to assume that we cannot access an ontology under a limited labeled data scenario.

\paragraph{Self-Supervised Learning}
There is a wide literature on self-supervision \cite{barlow1989unsupervised} and semi-supervised techniques \cite{chapelle2009semi}.
\citet{swayamdipta2018syntactic} introduce a syntactic scaffold, an approach to incorporate syntactic information into semantic tasks.
\citet{sankar-etal-2019-neural} found that Seq2Seq models are rarely sensitive to most perturbations, such as missing or reordering utterances. 
\citet{shi2019unsupervised} used variational RNN to extract latent dialogue structure and applied it to dialogue policy learning.
\citet{wu-etal-2019-self} introduced a self-supervised learning
task, inconsistent order detection, to explicitly capture the flow of conversation in dialogues. 
\citet{jin2018explicit} use unlabeled data to train probabilistic distributions over the vocabulary space as dialogue states for neural dialogue generation. 
\citet{su2020towards} provide both supervised and unsupervised learning algorithms to train language understanding and generation models in a dual learning setting.
\citet{tseng-etal-2019-semi} applied pseudo-labeling and $\prod$-model~\cite{sajjadi2016regularization} as additional semi-supervision to bootstrap state trackers. 
Our latent consistency comes from the consistency regularization~\cite{sajjadi2016regularization, berthelot2019mixmatch}, leveraging the idea that a model should output the same class distribution for an unlabeled example even after it has been augmented.



\section{Conclusion}
We investigate the potential of using self-supervised approaches for label-efficient DST in task-oriented dialogue systems. We strengthen latent consistency by augmenting data with stochastic word dropout and label guessing.
We model conversational behavior by the next response generation and turn utterance generation tasks. Experimental results show that we can significantly boost the joint goal accuracy with limited labeled data by exploiting self-supervision. 
We conduct comprehensive result analysis to cast light on and stimulate label-efficient DST.



\bibliography{acl2020}
\bibliographystyle{acl_natbib}

\clearpage
\newpage

\appendix
\input{Appendix}

\end{document}

%% file: Appendix.tex
\section{Appendices}
\label{sec:appendix}

\subsection{Data Statistics}
As shown in Table~\ref{DATASET-TABLE}, there are 30 \textit{(domain, slot)} pairs on MultiWOZ~\cite{multiwoz} dataset, including 10 slots in the \textit{hotel} domain, 6 slots in the \textit{train} domain, 3 slots in the \textit{attraction} domain, 7 slots in the \textit{restaurant} domain, and 4 slots in the \textit{taxi} domain. \textit{hotel} and \textit{restaurant} domains, in particular, have the most dialogues in the dataset. \textit{Taxi} domain, on the other hand, is the one with less data.


\begin{table}[ht]
\begin{center}
\resizebox{\linewidth}{!}{
\begin{tabular}{r|c|c|c|c|c}
\hline
 & \textbf{Hotel} & \textbf{Train} & \textbf{Attraction} & \textbf{Restaurant} & \textbf{Taxi} \\ \hline
Slots & \begin{tabular}[c]{@{}c@{}}price,\\ type,\\ parking,\\ stay,\\ day,\\ people,\\ area,\\ stars,\\ internet,\\ name\end{tabular} & \begin{tabular}[c]{@{}c@{}}destination,\\ departure,\\ day,\\ arrive by,\\ leave at,\\ people\end{tabular} & \begin{tabular}[c]{@{}c@{}}area,\\ name,\\ type\end{tabular} & \begin{tabular}[c]{@{}c@{}}food,\\ price,\\ area,\\ name,\\ time,\\ day,\\ people\end{tabular} & \begin{tabular}[c]{@{}c@{}}destination,\\ departure,\\ arrive by,\\ leave at\end{tabular} \\ \hline
Train & 3381 & 3103 & 2717 & 3813 & 1654 \\
Dev & 416 & 484 & 401 & 438 & 207 \\
Test & 394 & 494 & 395 & 437 & 195 \\ \hline
\end{tabular}
}
\end{center}
\caption{In total, there are 30 \textit{(domain, slot)} pairs from the selected five domains on MultiWOZ. The numbers in the last three rows indicate the number of dialogues.}
\label{DATASET-TABLE}
\end{table}

\begin{table*}
\centering
\resizebox{0.9\linewidth}{!}{
\begin{tabular}{r|r|c|c|c|c}
\hline
\multicolumn{2}{r|}{} & \textbf{1\%} & \textbf{5\%} & \textbf{10\%} & \textbf{25\%} \\ \hline
\multicolumn{2}{r|}{\textit{TRADE (w/o Ont.)}} & 10.35 (12.58) & 27.70 (31.17) & 32.61 (36.18) & 38.46 (42.71) \\ 
\hline
\multicolumn{2}{r|}{+ Consistency} & 13.77 (15.58) & 27.95 (31.89) & 33.64 (37.40) & 40.06 (43.45) \\ 
\hline
\multicolumn{2}{r|}{+ Behavior} & 18.69 (22.10) & 29.95 (34.30) & \textbf{34.57 (38.43)} & 39.42 (42.45) \\ 
\hline

\multicolumn{2}{r|}{Consistency + Behavior} & \textbf{19.50 (21.90)} & \textbf{30.59 (35.13)} & 34.50 (38.12) & \textbf{40.15 (43.40)} \\ 
\hline




\end{tabular}
}
\caption{Semi-supervised learning joint goal accuracy (and its fuzzy matching version in parentheses) on MultiWOZ-2.1 test set from 1\% to 25\% training data. }
\label{tb:trade_result_2.1}
\end{table*}




\begin{figure}[ht]
\centering
\begin{tabular}{|c|c|}
\hline
 & Dialogue History \\ \hline
\rotatebox[origin=l]{90}{100\% Data} & \includegraphics[width=0.82\linewidth]{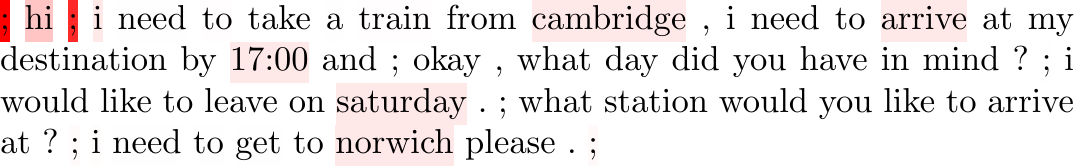} \\ \hline
\rotatebox[origin=l]{90}{1\% Data w/o SSL} & \includegraphics[width=0.82\linewidth]{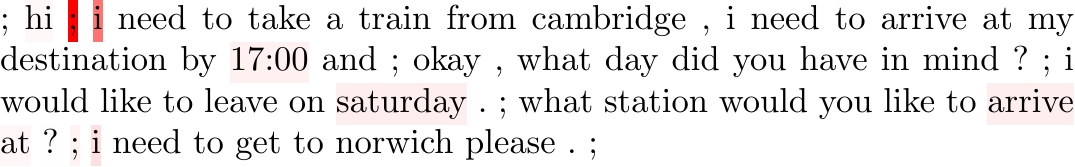} \\ \hline
\rotatebox[origin=l]{90}{1\% Data w/ SSL}  & \includegraphics[width=0.82\linewidth]{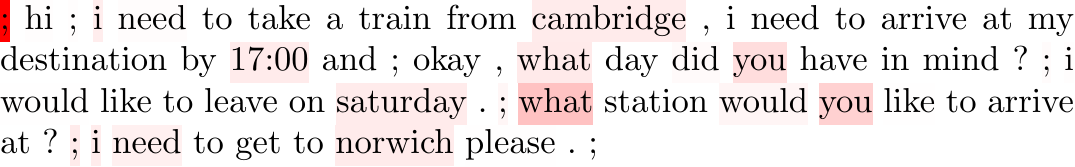} \\ \hline
\end{tabular}
\caption{Attention Visualization}
\label{attn_plot_appx3}
\end{figure}

\begin{figure}[ht]
\centering
\begin{tabular}{|c|c|}
\hline
 & Dialogue History \\ \hline
\rotatebox[origin=l]{90}{100\% Data} & \includegraphics[width=0.82\linewidth]{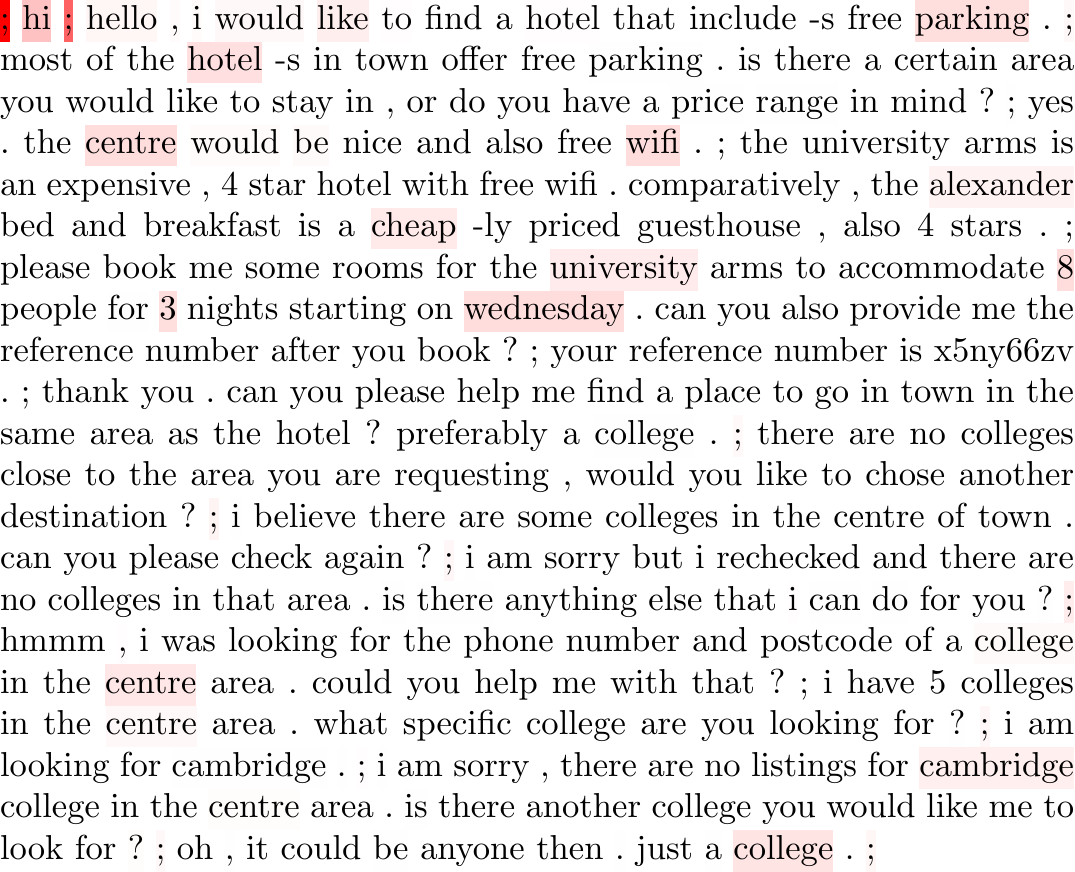} \\ \hline
\rotatebox[origin=l]{90}{1\% Data w/o SSL} & \includegraphics[width=0.82\linewidth]{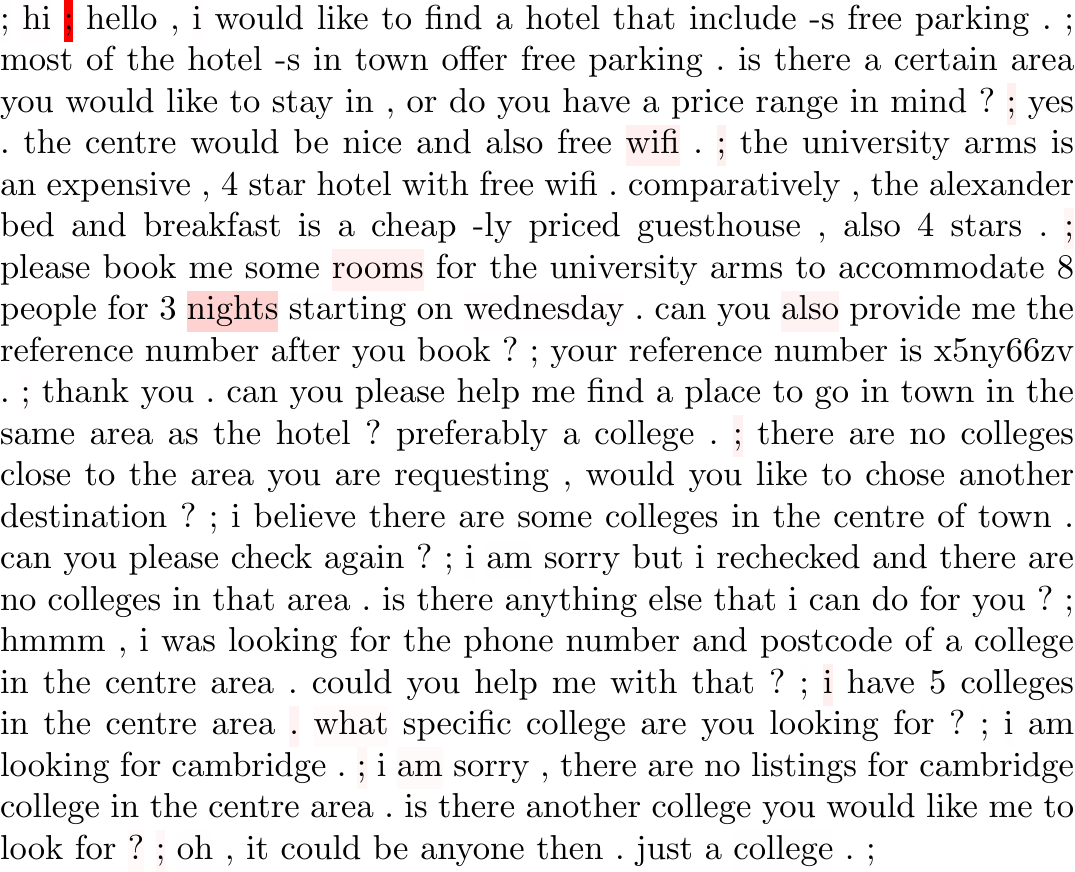} \\ \hline
\rotatebox[origin=l]{90}{1\% Data w/ SSL}  & \includegraphics[width=0.82\linewidth]{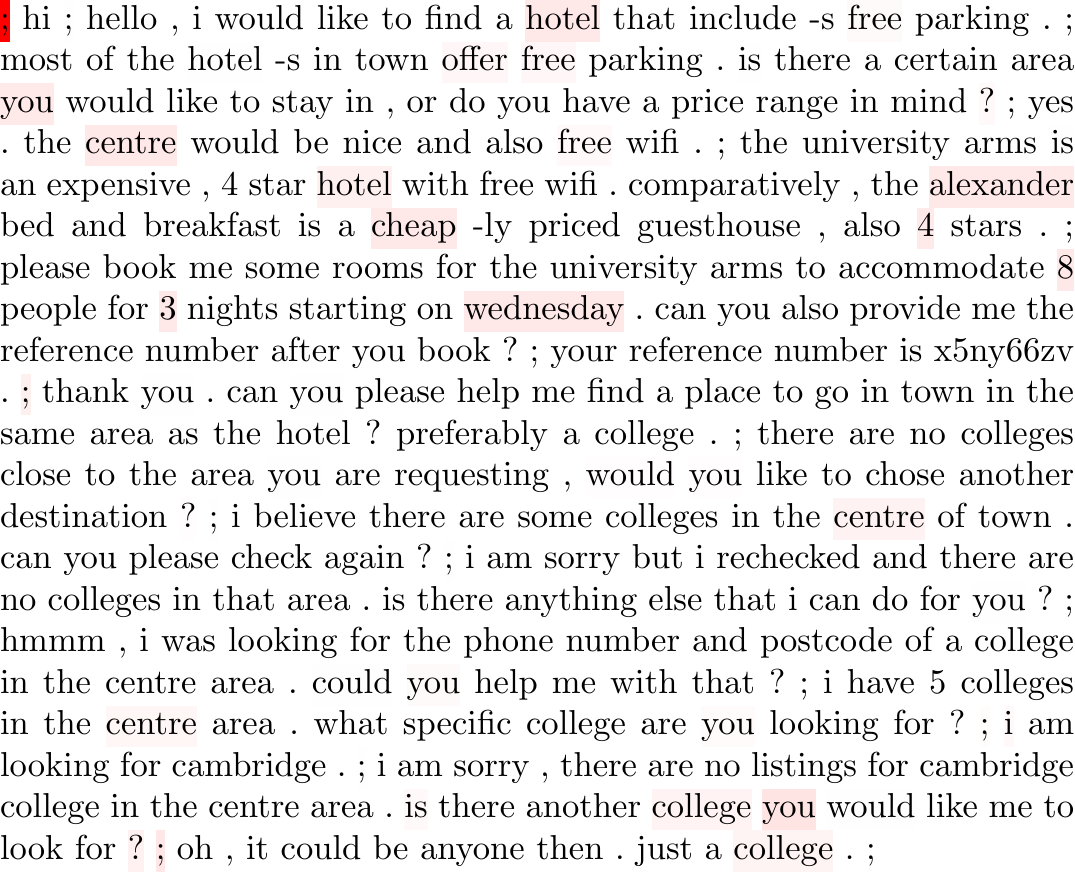} \\ \hline
\end{tabular}
\caption{Attention Visualization}
\label{attn_plot_appx2}
\end{figure}

\begin{figure}[ht]
\centering
\begin{tabular}{|c|c|}
\hline
 & Dialogue History \\ \hline
\rotatebox[origin=l]{90}{100\% Data} & \includegraphics[width=0.82\linewidth]{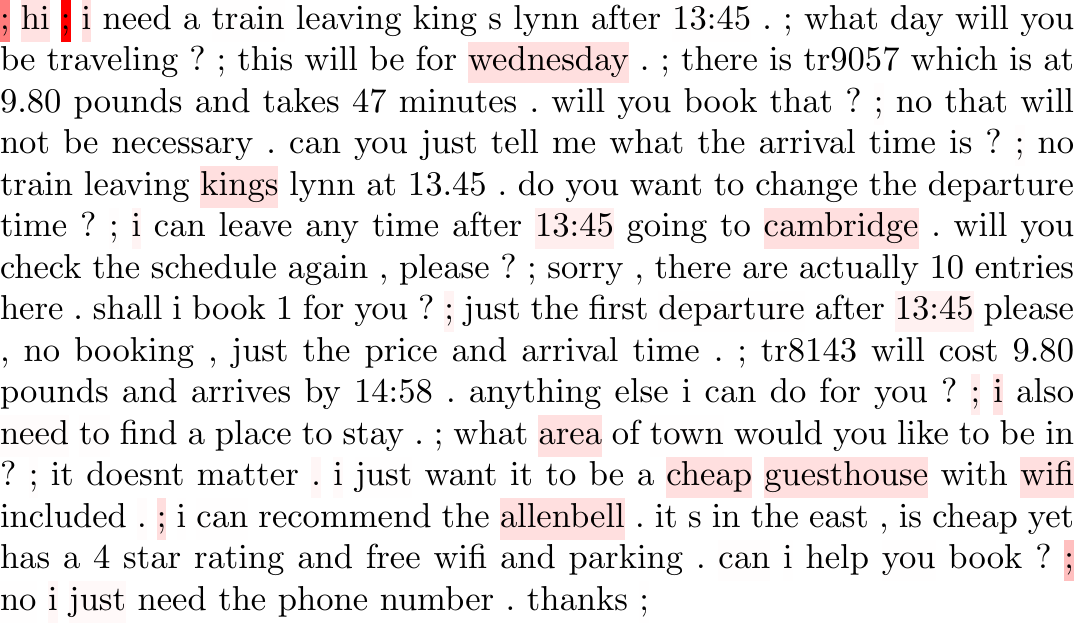} \\ \hline
\rotatebox[origin=l]{90}{1\% Data w/o SSL} & \includegraphics[width=0.82\linewidth]{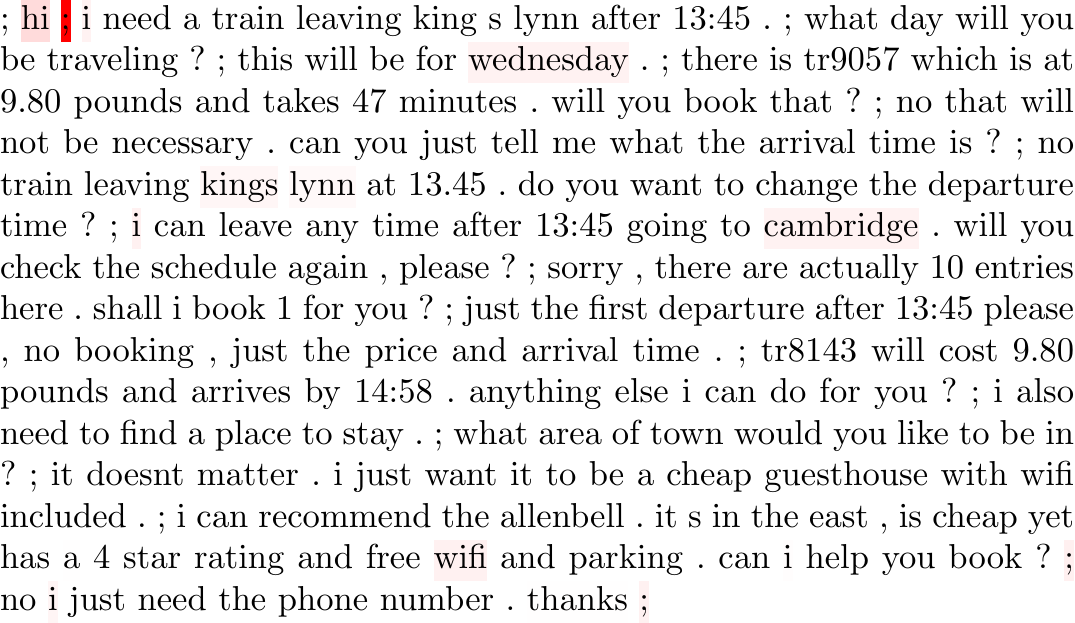} \\ \hline
\rotatebox[origin=l]{90}{1\% Data w/ SSL}  & \includegraphics[width=0.82\linewidth]{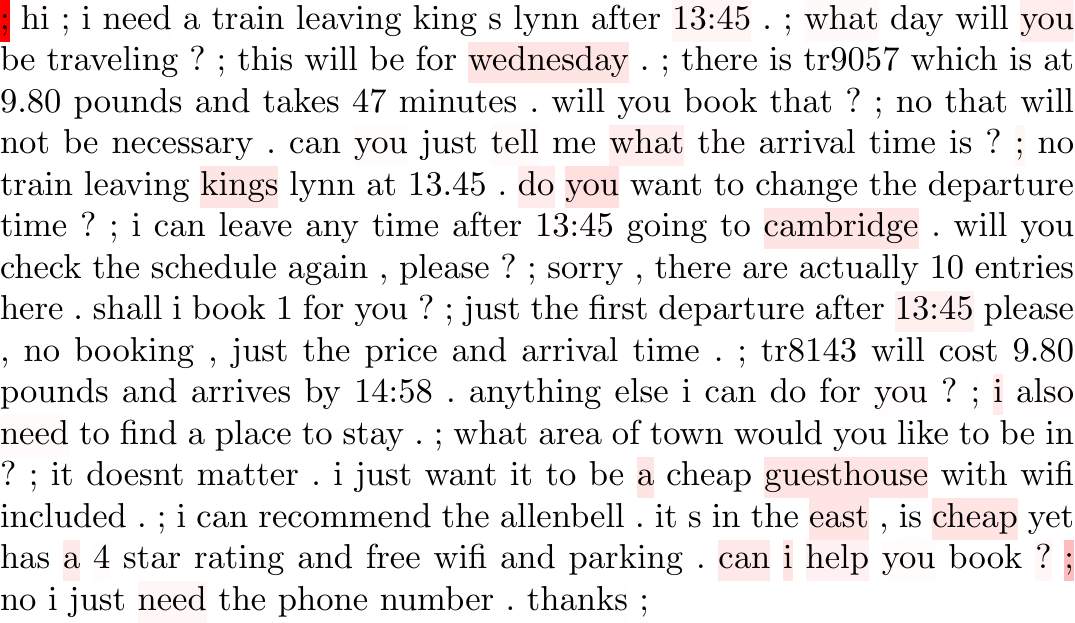} \\ \hline
\end{tabular}
\caption{Attention Visualization}
\label{attn_plot_appx1}
\end{figure}

\begin{figure}[ht]
\centering
\begin{tabular}{|c|c|}
\hline
 & Dialogue History \\ \hline
\rotatebox[origin=l]{90}{100\% Data} & \includegraphics[width=0.82\linewidth]{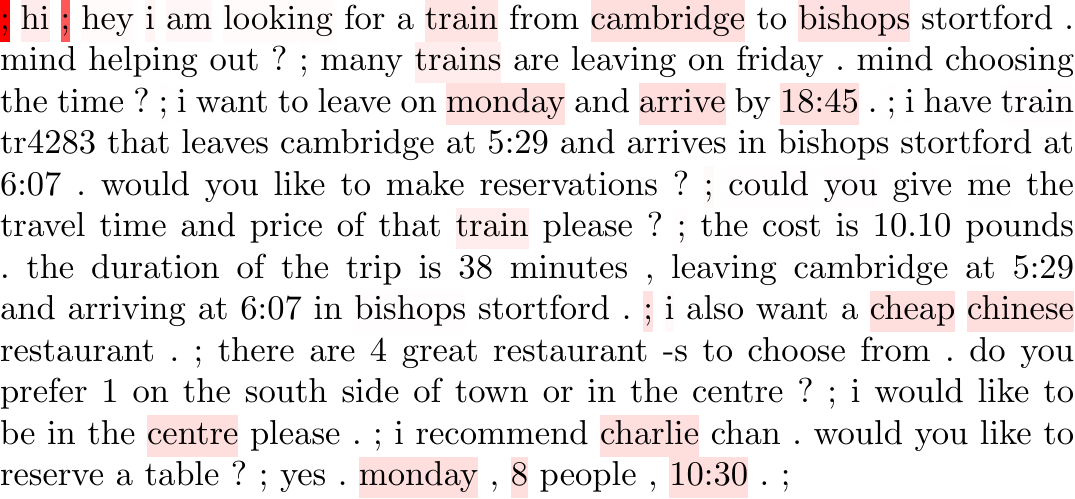} \\ \hline
\rotatebox[origin=l]{90}{1\% Data w/o SSL} & \includegraphics[width=0.82\linewidth]{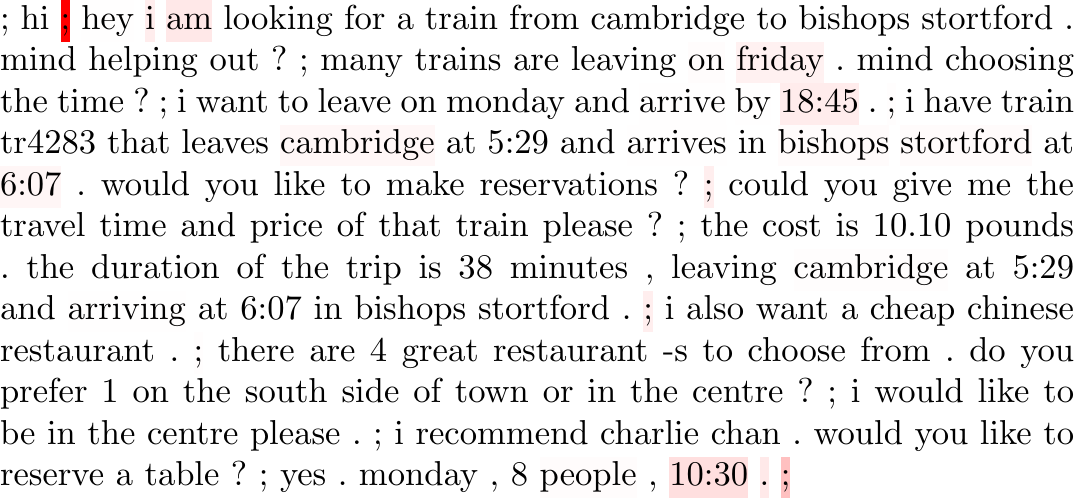} \\ \hline
\rotatebox[origin=l]{90}{1\% Data w/ SSL}  & \includegraphics[width=0.82\linewidth]{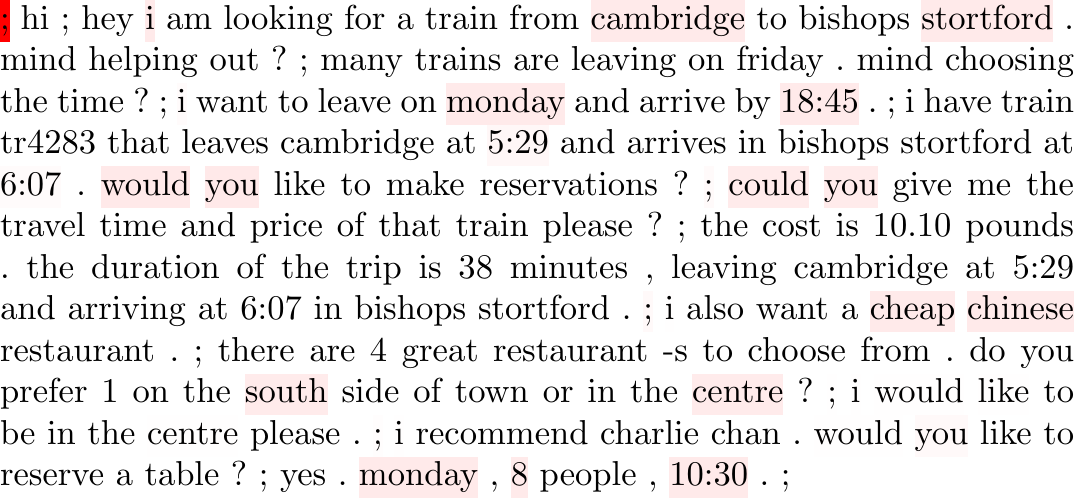} \\ \hline
\end{tabular}
\caption{Attention Visualization}
\label{attn_plot_appx4}
\end{figure}